\documentclass[a4paper, 10pt, conference]{ieeeconf}      
\usepackage{FG2025}
\usepackage{graphicx}
\usepackage{amsmath}
\usepackage[colorinlistoftodos]{todonotes}
\usepackage{booktabs}
\usepackage{multirow}
\usepackage{acronym}
\usepackage{hyperref}

\newcommand{\defineacronyms}{
  \acrodef{FPR}[FPR]{False Positive Rate}
  \acrodef{TPR}[TPR]{True Positive Rate}
}

\graphicspath{ {./images/} }
\FGfinalcopy 

\IEEEoverridecommandlockouts                              
\def\FGPaperID{****} 

\title{\LARGE \bf
JAM: A Comprehensive Model for Age Estimation, Verification, and Comparability
}

\author{\parbox{16cm}{\centering
    {\large François David$^{*,1}$, Alexey A. Novikov$^{*,1}$, Ruslan Parkhomenko$^2$, Artem Voronin$^1$, Alix Melchy$^1$}\\
    {\normalsize
    $^1$ Montreal, Canada, $^2$ Kyiv, Ukraine}\\
    {\normalsize $^*$These authors contributed equally to this work.}}
}

\begin{document}

\defineacronyms

\ifFGfinal
\thispagestyle{empty}
\pagestyle{empty}
\else
\pagestyle{plain}
\fi
\maketitle

\begin{abstract}
This paper introduces a comprehensive model for age estimation, verification, and comparability, offering a comprehensive solution for a wide range of applications. It employs advanced learning techniques to understand age distribution and uses confidence scores to create probabilistic age ranges, enhancing its ability to handle ambiguous cases. The model has been tested on both proprietary and public datasets and compared against one of the top-performing models in the field. Additionally, it has recently been evaluated by NIST as part of the FATE challenge, achieving top places in many categories.
\end{abstract}


\section{INTRODUCTION}

Age verification is crucial across industries due to regulatory demands and the need for secure interactions~\cite{ca2024designcode, la2024ageverification}. Ensuring users meet age requirements is essential for legal compliance, safeguarding minors, and preventing fraud. Our group has developed a robust model addressing three key aspects:

\begin{enumerate} 
\item \textbf{Age Estimation:} Predicting an individual's age from digital images (e.g., selfies). This allows platforms to provide age-appropriate content and comply with legal standards.

\item \textbf{Age Verification:} Comparing the estimated age against a required age threshold (e.g., 18, 21, 25) to grant or restrict access. This is vital for services offering age-restricted content. 
\item \textbf{Age Comparability:} Comparing the age from a selfie with the claimed age from an ID. While not common in general use, it's crucial for identity verification to detect fraud by ensuring the individual's appearance matches their claimed age.
\end{enumerate}

Our work considers all these dimensions to tailor solutions to our clients' diverse needs. We conducted extensive testing to ensure our model performs well across different demographic groups. 
This paper is organized as follows: In the Sec.~\ref{sec:related_work}, we highlight existing works on age estimation and outline the differences between those approaches and ours. In the Sec.~\ref{sec:methodology}, we discuss our custom loss function and its constituent components in detail. Finally in the Sec.~\ref{sec:results}, we describe the datasets and showcase evaluations on age estimation, age verification, and age comparability.

\section{RELATED WORK}
\label{sec:related_work}


Our work builds upon extensive research in age estimation, incorporating specialized models and advanced distribution learning techniques.


Adaptive Label Distribution Learning (ALDL) was introduced to provide sample-specific variances in facial age estimation~\cite{Geng2014FacialAE}. Pan et al.~\cite{Pan2018} proposed Mean-Variance Loss, foundational for distribution-based loss functions, though it treated the label distribution as an auxiliary component rather than integral to the model. Qiang et al.~\cite{Wen2020} refined the idea by applying Kullback-Leibler divergence to a discrete output vector, but this required an additional curated dataset and multiple forward passes per training iteration, reducing efficiency.

Integrating multimodal data has enhanced robustness in age estimation. MiVOLO-D1~\cite{Kuprashevich2023} and MiVOLO-V2~\cite{Kuprashevich2024} combined facial and full-person images, achieving state-of-the-art results. While MiVOLO-V2 improved performance with a larger dataset, gains were partly due to increased data volume rather than solely architectural innovations.
 


Our method aligns with advanced distribution learning techniques, building on prior ALDL approaches~\cite{Geng2014FacialAE, unimodal, Pan2018, Wen2020} to produce sample-specific variance for age estimation. Unlike previous methods that rely on categorical outputs with modality and distribution constraints, we simplify the process by directly outputting two continuous regression values. We introduce a balancing term that optimizes both the mean and variance simultaneously, reducing complexity and enhancing efficiency while delivering competitive performance across various age estimation tasks.

\section{METHODOLOGY}
\label{sec:methodology}

Our team has developed a novel, differentiable loss function to enhance age estimation models by incorporating confidence measures. This innovation enables the models to generalize effectively across a diverse range of selfies with varying quality and demographic characteristics. The loss function comprises three components, each with adjustable thresholds that can be fine-tuned based on specific contexts. Our model architecture outputs two values representing a probability distribution over the possible ages of the subjects. These outputs can be interpreted as the mean and standard deviation of a Gaussian distribution, which are used to assess prediction confidence. 


\subsection{Breakdown of the Loss Function Components}
The developed loss function is structured as follows, with detailed definitions provided for each term:

\begin{equation}
    L_{\text{jam}} = \alpha \, L_{\text{reg}} + \beta \, L_{\text{std}} + \delta \, L_{\text{dist}}
    \label{eq:jam_loss}
\end{equation}

The \textbf{regression} term $L_{\text{reg}}$ quantifies the average error and is critical as it guides the model to produce mean predictions as close as possible to the target values. The magnitude of this term is directly influenced by the discrepancy between the predicted age and the actual age. 

The \textbf{standard deviation} term $L_{\text{std}}$ serves as a penalty for larger standard deviations. This term encourages the network to generate increasingly smaller standard deviations, which is crucial for accurate predictions, especially for older age groups that are inherently more challenging to predict accurately.

The \textbf{distribution} term $L_{\text{dist}}$ plays a crucial role in balancing the regression and standard deviation terms. It penalizes the model significantly when the mean prediction deviates from the target, based on the standard deviation. This term ensures the model is precise in its predictions when the standard deviation is small and allows for larger deviations when there is greater uncertainty.

The \textbf{age decay $AD$} factor decreases as the age target increases. This behavior is instrumental in assigning greater significance to samples of younger ages, thereby prioritizing the acquisition of essential features for these target groups. 

Each of these loss components is defined as follows, incorporating the age decay factor which is exponentiated according to specific hyperparameters:


\begin{equation}
L_{\text{reg}} = \frac{1}{N} \sum_{i=1}^{N} \left| y_{\mu_i} - \hat{y}_i \right| \times AD_i^r
\end{equation}

\begin{equation}
L_{\text{std}} = \frac{1}{N} \sum_{i=1}^{N} y_{\sigma_i} \times AD_i^s
\end{equation}

\begin{equation}
L_{\text{dist}} = \frac{1}{N} \sum_{i=1}^{N} \left( \frac{y_{\mu_i} -  \hat{y}_i}{y_{\sigma_i} + c} \right)^2 \times AD_i^d
\end{equation}

The age decay factor $AD_i$ is defined as:
\begin{equation}
AD_i = \left(1 - \frac{\hat{y}_i}{M}\right)^2
\end{equation}

where $\hat{y}_i$ represents the actual age, $y_{\mu_i}$ and $y_{\sigma_i}$ are the mean and standard deviation predictions of the model, respectively. $\alpha$, $\beta$, $\delta$, and the exponents $r$, $s$, $d$, are tunable hyperparameters. 

The parameter \( M \) is termed the \textit{maximum age normalization factor}, serving to scale and normalize the predicted age values. This factor is essential for ensuring that the decay component is appropriately adjusted for the range of age values within the dataset, allowing for a consistent application across different age groups.

\subsection{Testing \& Inference}

After optimizing the model, the outputs $y_{\mu}$ and $y_{\sigma}$ can be used to gauge the confidence range for estimating a subject’s age. For the sole task of age estimation, we rely on $y_{\mu}$ as the predicted age. When the model is well-trained, it will produce smaller distributions when confident and larger ones when less certain. This is crucial for age comparability and verification tasks. To precisely scale the model's outputs to a specific leniency level, each $y_{\sigma}$ is combined with data-defined thresholds based on age groups.

Using the model outputs and defined thresholds, we establish a confidence interval for age estimation. This interval is defined by the following range:
\begin{equation}
    [y_{\mu} - y_{\sigma} \times \text{LT}_j, \, y_{\mu} + y_{\sigma} \times \text{UT}_j]
\label{eq:ranges}
\end{equation}

This range defines the lower and upper bounds of the estimated age, ensuring that predictions are tailored to specific confidence levels determined by the $\text{LT}_j$ (Lower Threshold) and $\text{UT}_j$ (Upper Threshold). These thresholds are calibrated against a supplementary validation set that mirrors production distributions across specific age group buckets $j$ to achieve targeted \ac{FPR}.

The age group buckets are typically organized in 5-year intervals, such as a [20, 25) bucket, where all predictions $y_{\mu}$ falling within that range would use the associated $\text{LT}_j$ and $\text{UT}_j$  thresholds. This method enables more dynamic, confidence-aware age estimation, leading to more accurate and reliable outcomes across various demographics and image qualities. For age verification or comparability, one can simply use the age ranges described above, fine-tuned to their specific use case. If the goal is to verify that a person is above a certain age $T$, the focus would shift to calculating the age range and ensure that the lower limit of the range exceeds $T$.

In theory, the output distribution can be modeled as a function composed of two normal distributions, both sharing the same mean. This results in a probability distribution function configured as a piecewise function with the following parameters:
\begin{equation}
P(x; y_{\mu}, y_{\sigma}) =
\begin{cases}
    \mathcal{N}(y_{\mu}, (y_{\sigma} \times \text{LT}_j)^2) & \text{if } x < y_{\mu}, \\
    \mathcal{N}(y_{\mu}, (y_{\sigma} \times \text{UT}_j)^2) & \text{if } x \geq y_{\mu}.
\end{cases}
\end{equation}

\begin{figure}[!ht]

\centering
\includegraphics[width=0.3\textwidth]{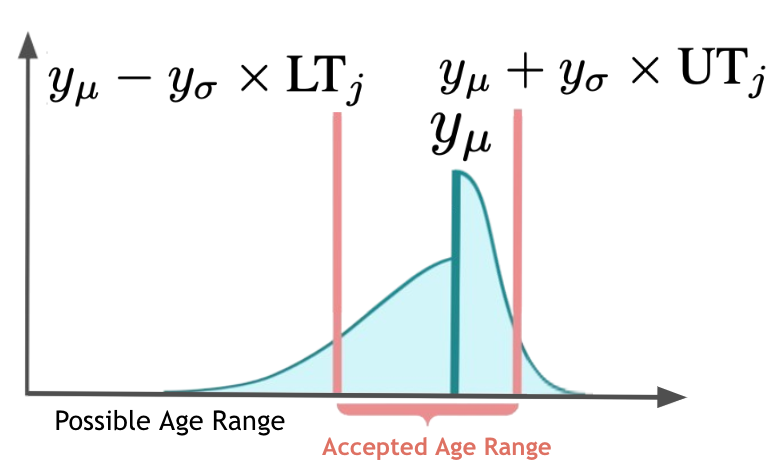}
\caption{A visual of the model's age probability functions and thresholds shows significant shifts with varying strictness. The pink area marks the accepted age range for a hypothetical age comparability task.}
\end{figure}

\section{RESULTS AND DISCUSSION}
\label{sec:results}

\subsection{Setup}
In our experiments, we utilized ResNet50 as the backbone architecture. Training was exclusively conducted using data from our live traffic, for which explicit user consent was obtained prior to training. We maintain rigorous privacy, security, and data retention policies for both source data, such as selfies, and biometric data. 

During initial experiments, we fine-tuned the model's parameters, ultimately setting the loss function parameters as follows: $\alpha = 1$, $\beta = 1$, $\theta = 1.5$, $\lambda_r = 1$, $\lambda_s = 1.5$, and $\lambda_d = 2$. We also set the maximum age normalization factor \( M \) to 115. 

\subsection{Datasets}

We evaluated our model on two datasets to ensure robust testing across various conditions. The first dataset, referred to as the \textbf{Proprietary Dataset (JPD)}, consists of 113,000 selfies from our customers and represents a diverse global demographic with ages ranging from 3 to 91 years. Crafted to resemble production traffic, this dataset is intentionally made more challenging to more pronouncedly reveal differences between candidate models, aiding in the selection of the best performers. It includes adjustments for enhanced balance across ages, genders, and geographies, ensuring a comprehensive and fair evaluation environment.

The second dataset, the publicly available \textbf{ONOT}~\cite{di2024onot}, was selected due to its closer alignment with our operational environment compared to other public datasets.

Our model was also recently evaluated in the \textbf{NIST FATE} Age Estimation and Verification Challenge using various datasets~\cite{nist2024fateweb}. We summarize key points of the evaluation in Sec.~\ref{sec:nist_testing}.

For the training dataset, we used a dataset of the order of magnitude of $10^6$ selfies from our live traffic. Although it is not within the scope of this submission, we also conducted a small study on how dataset volume impacts performance. Increasing the dataset size by 40\% reduced the overall MAE by approximately 0.6, which corresponds to an error reduction of slightly over half a year.

\subsection{Age Estimation Results}

We evaluate our model's age prediction accuracy using the Mean Absolute Error (MAE), which measures the average absolute difference between predicted and actual ages.

We conducted a comparative analysis between our JAM model and the MiVolo model, recognized for its strong performance on several public benchmarks. This evaluation used both our proprietary JPD dataset and the publicly available synthetic dataset ONOT (see Table \ref{tab:performance_comparison_mivolo}). 

\begin{figure}
\centering
\label{fig:mae_comparison_chart}
\includegraphics[width=0.4\textwidth]{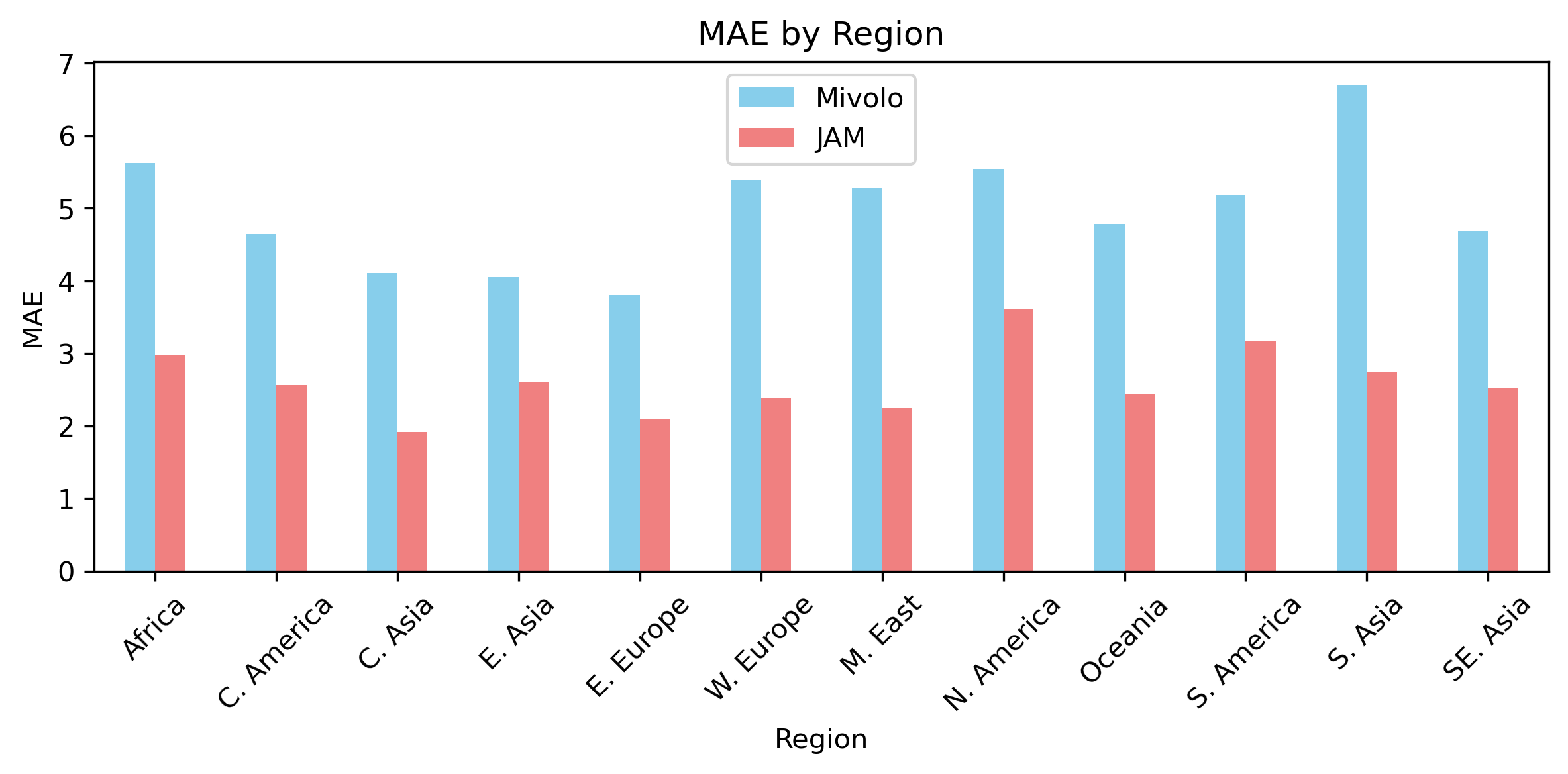}
\caption{Comparison of MAE scores between our model and the MiVOLO model across various regions. We used a regional breakdown similar to that in the NIST evaluations but included more regions}
\end{figure}

Our JAM model outperforms the MiVolo model on the JPD dataset overall and across all tested regions (Fig.~\ref{fig:mae_comparison_chart}).  This superior performance may be attributed to our distribution-based approach, which better handles the variabilities encountered in real user data. By modeling age predictions as probability distributions, JAM effectively captures the nuances and inconsistencies present in production environments, leading to more accurate and reliable age estimations.

Conversely, the minimal difference in performance on the ONOT dataset—which consists of synthetic mugshot images generated by stable diffusion—highlights that both our JAM model and the MiVolo model perform consistently when dealing with unfamiliar and non-realistic data scenarios. This close performance gap indicates that both models maintain robustness even when applied to synthetic datasets where the depicted ages are artificially assigned and do not correspond to real human aging patterns.

\begin{table}[h]
\caption{Comparative Performance of the Proposed JAM Model and Mivolo Architecture Across Two Datasets.}
\label{tab:performance_comparison_mivolo}
\begin{center}
\begin{tabular}{|c|c|c|}
\hline
\multicolumn{3}{|c|}{Mean Average Error by Dataset} \\
\hline
& \textbf{JAM} & \textbf{MiVolo } \\
\hline
\textbf{JPD} & 2.82 & 5.18 \\
\hline
\textbf{ONOT} & 11.24 & 11.04 \\
\hline
\end{tabular}
\end{center}
\end{table}

\subsection{Age Verification Results}

\begin{table}[h]
\caption{Comparison of Performance of the Confidence and Single Value Age Verification}
\label{tab:age_verification}
\begin{center}
\begin{tabular}{|c|c|c|}
\hline
\multicolumn{3}{|c|}{JPD/ Age Verification of 18+ / Challenge-25}\\
\hline  \hline
Range Method &  \acl{FPR} &  \acl{TPR}\\
\hline
JAM Confidence & \textbf{0.001} & \multirow{2}{*}{0.8142} \\
\cline{1-2}
JAM SR & 0.007 & \\
\hline\hline
\multicolumn{3}{|c|}{ONOT / Age Verification of 21+ / Challenge-28}\\ 
\hline 
Range Method &  \acl{FPR} &  \acl{TPR}\\ 
\hline
JAM Confidence & \textbf{0.0} & \multirow{2}{*}{0.7568} \\
\cline{1-2}
JAM SR & 0.008 & \\
\hline
\end{tabular}
\end{center}
\end{table}

In this evaluation, we compared two methods for performing age verification tasks, inspired by the well-established Challenge-25 and Challenge-28 policies used by retailers of age-restricted products. The Challenge-25 policy requires customers who appear under 25 to present identification to confirm they are at least 18 years old. Similarly, the Challenge-28 policy requires anyone who looks under 28 to show identification to confirm they are at least 21. Since the ONOT dataset contains no subjects under 18, we applied the Challenge-28 policy for this dataset.

The first method, referred to as JAM Singular Regression (SR), employs a traditional approach where a single predicted age value is compared against an age threshold with an added safety buffer (25 years for Challenge-25 and 28 years for Challenge-28). In the JPD dataset (Challenge-25), this method achieved an \ac{FPR} of 0.007 (0.7\%) and a \ac{TPR} of 81.42\%. In the ONOT dataset (Challenge-28), it achieved an \ac{FPR} of 0.008 (0.8\%) and a \ac{TPR} of 75.68\%.

The second method is a confidence-based approach that utilizes the distributional output of our model. By ensuring that the lower bound of the user's predicted age range is above the legal age limit, we make more informed decisions based on the model's confidence. Adjusting the threshold leniency to match the \ac{TPR} of the SR method, the confidence-based approach significantly reduced the \ac{FPR} to 0.001 (0.1\%) for the JPD dataset and to 0\% for the ONOT dataset. This represents a sevenfold reduction in false positives for the former and a complete elimination of false positives for the latter.

These results demonstrate that leveraging the model's confidence leads to a more effective age verification process. The confidence-based method reduces the likelihood of incorrectly flagging adults as underage (\ac{FPR}), while maintaining the same level of detection for actual underage individuals (\ac{TPR}). 

\subsection{Age Comparability Results}

Age comparability is not commonly addressed in general applications but is particularly important in the identity verification space.

\begin{figure}[h]
\centering
\includegraphics[width=0.35\textwidth]{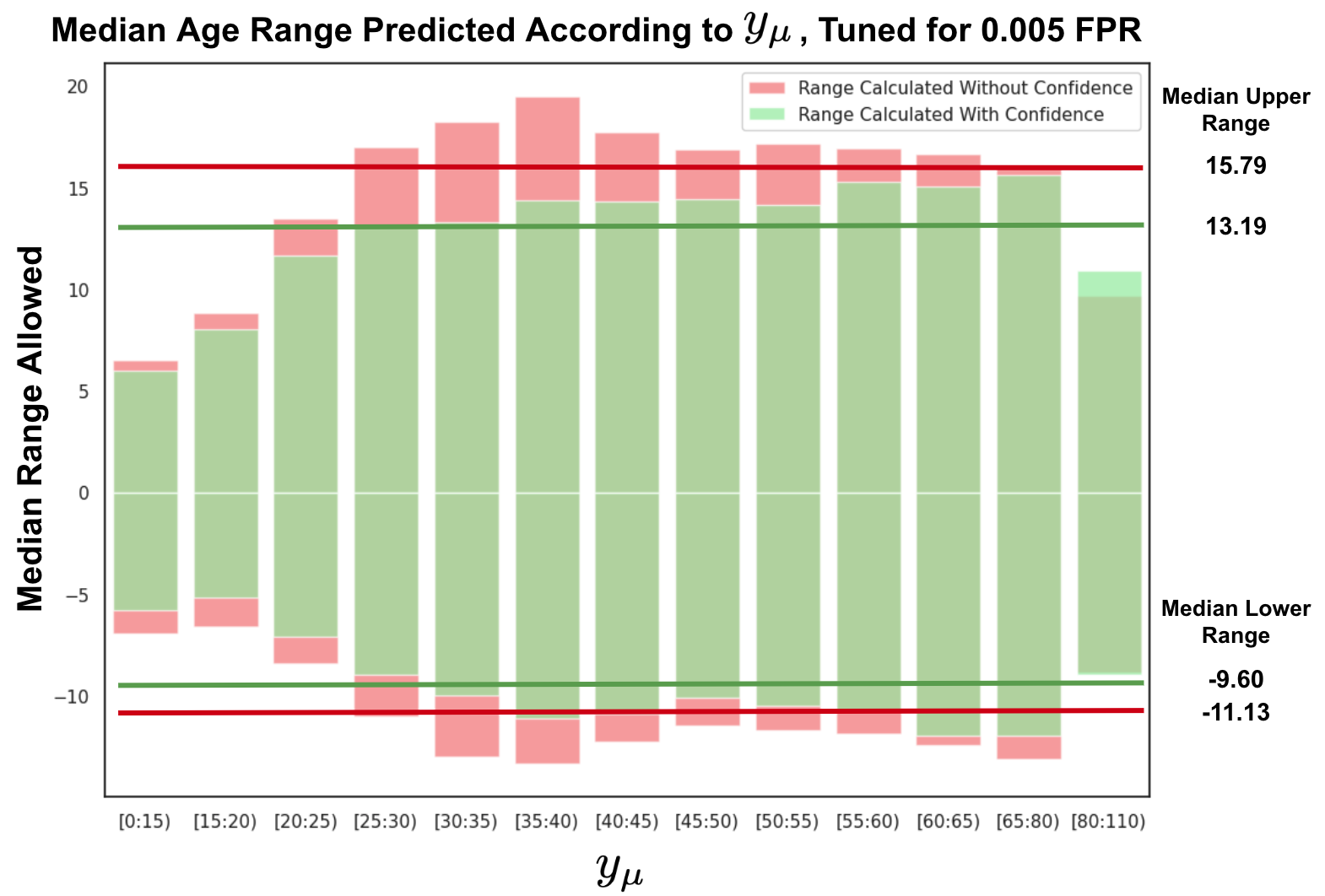}
\caption{Visual representation of the age comparability ranges obtained with and without taking into consideration the confidence. Tuned for 0.5\% \ac{FPR} on the JPD Dataset. }
\label{fig:age_comparability}
\end{figure}

As shown in Fig.~\ref{fig:age_comparability}, incorporating $y_{\sigma}$ confidence into the decision-making process when predicting a person's age range can significantly reduce that range, thereby increasing the precision of the prediction while maintaining the same \ac{FPR}. In this analysis, we subdivided the dataset into smaller subsets based on where $y_{\mu}$ (mean prediction) fell. For each subset, we calculated the appropriate upper and lower bounds needed to achieve a specific \ac{FPR}.

For the method incorporating $y_{\sigma}$, we selected thresholds that ensured the range, as explained in Eq.~\ref{eq:ranges}, met the \ac{FPR} criterion, which we set at 0.5\%. In contrast, the method without $y_{\sigma}$ simply calculated the usual upper and lower limits without considering the model's confidence. This approach tends to be suboptimal, as it disregards the model's confidence level and can be disproportionately affected by low-quality or uninformative images, such as when a user is wearing a mask.

In this experiment, thanks to the reliable confidence output from the model, we were able to reduce the median predicted age range from 26.92 to 22.80 on the JPD dataset at a fixed \ac{FPR} of 0.5\%. This represents a reduction of approximately 15.3\%. In the context of identity verification, this narrower age range enhances the detection of identity fraud, leading to a safer and more reliable system overall.

\subsection{NIST testing}
\label{sec:nist_testing}

Our age estimation model has been independently evaluated by the National Institute of Standards and Technology (NIST), affirming its competitive scores in operational datasets, particularly Application and Mugshot. 

In the full NIST report \cite{nist2024fatepdf}, Table 13 provides a detailed overview of age estimation accuracy by sex, mean absolute error (MAE), across different age bands. For a detailed breakdown of our model's performance, please see Table \ref{tab:performance_summary} in this document, which presents comprehensive results across both Mugshot and Application datasets.

\begin{table}[ht]
\centering
\caption{Performance Summary in NIST FATE Challenge}
\label{tab:performance_summary}
\begin{tabular}{@{}llll@{}}
\toprule
\textbf{Category}            & \textbf{Age Group} & \textbf{Placement}            & \textbf{Gap to 1st}           \\ \midrule
\multicolumn{4}{c}{\textbf{Mugshot Dataset}}                                      \\ \addlinespace
First Place                  & Female 51-80       & 1st                          & -                              \\
Second Place                 & Male 51-80         & 2nd                          & approx. 1 month               \\
                             & Male 31-50         & 2nd                          & approx. 1 month               \\
                             & Female 31-50       & 2nd                          & approx. 1 month               \\
Third Place                  & Male 18-30         & 3rd                          & approx. 4 months              \\
Seventh Place                & Female 18-30       & 7th                          & approx. 1 year                \\ \addlinespace
\midrule
\multicolumn{4}{c}{\textbf{Application Dataset}}                                  \\ \addlinespace
First Place                  & Male 51-80         & 1st                          & -                              \\
                             & Female 31-50       & 1st                          & -                              \\
Second Place                 & Female 51-80       & 2nd                          & approx. 1 month               \\
                             & Male 31-50         & 2nd                          & approx. 1 month               \\
                             & Male 18-31         & 2nd                          & approx. 1 month               \\
Fourth Place                 & Female 18-30       & 4th                          & approx. 6 months              \\ \bottomrule
\end{tabular}
\end{table}

Our model also secured the first place in the inconvenience score across all threshold (T) settings, as shown in Table 8 of the report. This result demonstrates robust performance of our model in processing scenarios without compromising user convenience.


\section{CONCLUSION}
\label{sec:conclusion}

This paper highlights our advanced model primarily designed for age estimation, with applicability to other tasks such as verification and comparability, offering a robust alternative to existing methods. Utilizing innovative techniques and a specialized loss function, our model meets and often exceeds conventional standards, providing practical benefits for precise age range predictions. It effectively handles real-world challenges, including diverse demographics and varying image qualities, making it ideal for age verification applications. The model’s confidence-based output ensures accuracy under diverse conditions, thereby enhancing user safety. Furthermore, it consistently achieves competitive results on both public and proprietary datasets, and has shown strong performance in recent NIST FATE challenge evaluations.


\section*{ETHICAL IMPACT STATEMENT}

\subsection{Risk}

We have not sought an external ethical review board for the age estimation and verification aspects of this project, as both areas are well-established in the industry and are widely recognized as low-risk from an ethical standpoint. Age estimation and verification is generally more accurately performed by modern ML/AI systems than by humans. AWS Rekognition demonstrates superior accuracy to humans in 9 out of 12 demographic categories~\cite{Ganel2022}. 

The project and the data have been reviewed internally by our company's ethics council and evaluated against our ethical policies. No violations of laws or ethical concerns have been identified.
One aspect to assess from a risk perspective is potential bias, as many age estimation systems exhibit bias issues. For example, age-related biases are a common concern in artificial intelligence systems.

One aspect to assess from a risk perspective is potential bias, as many age estimation systems exhibit bias issues. For example, age-related biases are a common concern in artificial intelligence systems~\cite{Chu2023}.

\subsection{Mitigation}

To mitigate bias, we employed curated datasets for model training and testing. These datasets are balanced across various age groups, genders, and countries of origin. However, as observed in NIST reports~\cite{nist2024fatepdf}, our solution still shows some biases—for instance, age estimation tends to be more accurate for males than for females (a trend observed across all participants and age groups), and middle-aged individuals (20–50 years) are more accurately assessed than older populations.

We hypothesize that age estimation accuracy may decrease in older populations because age-related changes in appearance become less pronounced, and gender bias could be influenced by cultural factors such as the use of cosmetics.

To address concerns regarding inappropriate use and to ensure transparency and accountability, our approach includes strict usage guidelines and the publication of detailed metrics that quantify the risks associated with model errors. These metrics are designed to provide clear insights into the system's performance across different demographics and scenarios, enabling continuous monitoring and improvement.

In our live traffic, we apply different mitigation strategies depending on the specific application:

\begin{itemize} 
    \item \textbf{Age Comparability}: Large age deltas are applied when making decisions to reject users in fraud detection applications. Such systems only act when a significant age mismatch is detected. 
    \item \textbf{Age Estimation}: Large confidence intervals are used, and age values are not directly applied or utilized in real applications due to the risk of errors. 
    \item \textbf{Age Verification}: Both large deltas and confidence intervals can be utilized. For instance, if a user attempts to access a restricted content website, the verification system rejects the users only when it is highly certain that the user is underage. 
\end{itemize}

\subsection{Privacy and Security}

For all real images of individuals used in the research, we obtained direct consent from each user. When users utilize our applications for identity verification, we ask for their consent to use their data for commercial R\&D applications, including the development of age models. 

There is no direct compensation provided to the end user, but they benefit from improved service after certain development iterations. Only data from users above 18 years of age have been used, as evidenced by the government-issued ID documents associated with the data.

The image data is stored securely and encrypted, with eventual deletion in accordance with the retention policy or under data subject rights requests. No information or images that could lead to user identification are included in this manuscript. 

Additionally, the age data obtained from AI models is not stored and is removed immediately after analysis, ensuring that no sensitive information is retained.

\newpage
\bibliography{egbib}
\bibliographystyle{unsrt}

\end{document}